# ODVICE: An Ontology-Driven Visual Analytic Tool for Interactive Cohort Extraction

Mohamed Ghalwash
mohamed.ghalwash@ibm.com
IBM T. J. Watson Research Center
Yorktown Heights, NY, USA

Zijun Yao
zijun.yao@ibm.com
IBM T. J. Watson Research Center
Yorktown Heights, NY, USA

Prithwish Chakrabotry
prithwish.chakraborty@ibm.com
IBM T. J. Watson Research Center
Yorktown Heights, NY, USA

James Codella
jvcodell@us.ibm.com
IBM T. J. Watson Research Center
Yorktown Heights, NY, USA

Daby Sow
sowdaby@us.ibm.com
IBM T. J. Watson Research Center
Yorktown Heights, NY, USA

## ABSTRACT

*Increased availability of electronic health records (EHR) has enabled researchers to study various medical questions. Cohort selection for the hypothesis under investigation is one of the main consideration for EHR analysis. For uncommon diseases, cohorts extracted from EHRs contain very limited number of records – hampering the robustness of any analysis. Data augmentation methods have been successfully applied in other domains to address this issue mainly using simulated records. In this paper, we present ODVICE, a data augmentation framework that leverages the medical concept ontology to systematically augment records using a novel ontologically guided Monte-Carlo graph spanning algorithm. The tool allows end users to specify a small set of interactive controls to control the augmentation process. We analyze the importance of ODVICE by conducting studies on MIMIC-III dataset for two learning tasks. Our results demonstrate the predictive performance of ODVICE augmented cohorts, showing ∼ 30% improvement in area under the curve (AUC) over the non-augmented dataset and other data augmentation strategies.*

## INTRODUCTION

Increasing adoption of electronic data collection systems for healthcare is leading to an ever-growing availability of rich, longitudinal records spanning millions of patients and capturing significant portion of their medical histories. While these data sets typically contain large number of patients, constructing cohorts for patients with specific conditions brings this number down radically, often to numbers that are too small for the application of modern techniques.

Data augmentation has been used to tackle data scarcity problems in various tasks such as Natural Language Processing [10] and image classification [9, 12, 14]. However, many of these methods revolve around the notion of being able to *synthetically* generate data by applying various forms of noise or perturbations on the small amount of existing data. Analytical models [8], clinical workflows [15], and deep learning approaches, such as Generative Adversarial Networks (GANs) have shown some promise in generating synthetic EHR data[1, 2] and medical images [5, 13]. These methods require external validation, have a limited ability to model rare events, and may be a source of additional bias. Hence, it is desirable to use *real* data for augmentation when possible.

There have been a few notable approaches where real data have been used to augment data. For example, images from an external medical data set were used for augmentation by computing a similarity metric between the images and then selecting the top *k* most similar images for each image class for inclusion [16]. The utility of a public data set designed for generalized skin segmentation was demonstrated to augment the clinical data for segmenting healthy and diseased skin and achieve improved accuracy[3]. Patient similarity[11, 17] metrics can also be used to search for patients similar to the original cohort based on patients' attributes. These methods often lack to take into account the semantic relationship between diseases the patients have.

In this paper, we expand on the idea of using real data by augmenting data from patient records that may not belong in the original cohort obtained from a given user query. Beyond the use of a notion of similarity applied for data augmentation, we propose to use existing medical ontologies to guide the selection of these real data points outside of the original queried corpus that we believe to be not only close to this original corpus but also semantically related.

The contribution of our work is fourfold. First, we develop novel algorithms for filtering, sampling, and augmenting a subset of EHR data for a particular cohort using data from real patients found elsewhere in the EHR. Second, we leverage the graph-structure of the given medical ontology to identify data from patients that are semantically similar. Third, we implement a web-based interface to empower data scientists to quickly and easily navigate the graphical relationship of the data, obtain insights related to data quantity and diversity, and extract a subgraph and its associated data for use in the augmentation. Fourth, we demonstrate the utility of the *real* augmented data to boost the performance of analytic models using logistic regression and random forest as show cases for predicting in-hospital mortalities and acute myocardial infarction for rheumatic disease of heart valve or mitral valve stenosis patients.

## METHODS

We present the proposed ODVICE framework for data augmentation. ODVICE uses a medical concept ontology to guide the cohort augmentation process. The framework is agnostic to the ontology being used; however, in the following we use SNOMED CT ontology which is recognized as the recommended clinical reference terminology for use in clinical information systems [4]. SNOMED CT can be depicted as a graph of codes representing clinical terminology used in electronic health records. In a nutshell, ODVICE leverages SNOMED CT for data augmentation by discovering medical concepts that are semantically related to the AI task in-hand, and using cohorts of those relevant concepts to expand any initially pre-selected patient cohort. For example, given a task of detecting the myocardial infraction phenotype for patients that have one of the following disorders: mitral valve stenosis (SNOMED CT code 79619009) or rheumatic disease of the heart valve (SNOMED CT code 16063004), ODVICE allows the user to expand its data set with patients with conditions that are related to heart diseases by leveraging the graph structure of the SNOMED CT ontology.

Effectively, we represent the knowledge graph of SNOMED CT with a directed acyclic graph with nodes corresponding to medical concepts and edges corresponding to ontology relationships. Let $\mathcal{S}$ denote this graph and let $\mathcal{D}$ denote the EHR data set that we have access to for AI modeling purposes. Using $\mathcal{S}$ and $\mathcal{D}$, we construct $\mathcal{G}$ as a subgraph of $\mathcal{S}$ with nodes corresponding to SNOMED CT codes that are present in $\mathcal{D}$. At each node $n \in \mathcal{G}$, ODVICE tracks the following information:

- $c_n$, the SNOMED code for the node $n$ given obtained from $\mathcal{S}$.
- $V_n$, the set of all visits $v \in \mathcal{D}$ corresponding to $c_n$.
- $p_n$, empirical distribution of phenotype occurrences based on all $v \in V_n$, according to a pre-defined set of phenotypes $P$.

ODVICE takes user inputs to perform the following two operations:

(1) **Data Filtering:** ODVICE generates an initial filtered graph $\mathcal{G}_F$ based on user inputs consisting of SNOMED CT codes and phenotypes of interest. This step constrains the data augmentation search space based on user specified interests.
(2) **Data Augmentation:** ODVICE generates an augmented graph $\mathcal{G}_A$ within $\mathcal{G}_F$ according to a sampling procedure that benefits from the underlying graph structure of $\mathcal{G}_F$ inherited from the original SNOMED CT graph $\mathcal{S}$.

### Data Filtering

ODVICE captures the intent of the end user to filter out nodes from $\mathcal{G}$ that are deemed not relevant for the analysis. This step constrains the analysis to nodes from $\mathcal{G}$ that explicitly contain patient visits with one of the user requested SNOMED CT codes and phenotypes.

Besides $\mathcal{G}$, the user specified disjunctive[1] list of SNOMED CT codes $S_u$ and the user specified disjunctive list of phenotypes $P_u$, the filtering operation also takes two thresholds $\eta_u$ and $\gamma_u^p$ controlling respectively the minimal amount of visits by nodes and by specific phenotypes that each node of the filtered graph is required to have. ODVICE iteratively considers every node $n \in \mathcal{G}$ if it has enough visits for analysis (i.e with the size of $V_n$ greater than $\eta_u$) and if at least one of the phenotypes of interest is well represented (i.e. with an occurrence greater than $\gamma_u^p$). Based on these two conditions, nodes and their descendants are selected. We include descendants in the filtered graph since SNOMED CT guarantees that these nodes contain the medical concepts that specialize the medical concepts of their parents. With these steps, we are constraining our analysis to weakly connected components of $\mathcal{G}$ containing $S_u$ and satisfying all user specified visit & phenotype criteria. The output of this step is a filtered graph denoted $\mathcal{G}_F$.

---
[1] While the method described in this paper assumes that end user is specifying disjunctions (OR) on SNOMED CT codes and phenotypes, it can be extended to conjunctions and more complex queries.

### Data Augmentation

This method takes as inputs the filtered graph $\mathcal{G}_F$ together with user selected nodes $S_u$. It applies Algorithm 1 to grow $S_u$ within $\mathcal{G}_F$ by first creating the sub-graph from $S_u$ within $\mathcal{G}_F$ that we denote $\mathcal{G}_F^{S_u}$, and then collects nodes in $\mathcal{G}_F$ that have proximity and are similar to the nodes of $\mathcal{G}_F^{S_u}$. The proximity among nodes is determined by the graph structure of $\mathcal{G}_F$. Similarity among nodes is determine by computing the KL divergence between the empirical distribution of phenotypes of candidate nodes in $\mathcal{G}_F^{S_u}$ and nodes in $S_u$. More specifically, ODVICE discovers a candidate node $q \in \mathcal{G}_F^{S_u}$ to augment $S_u$ if $\forall n \in \mathcal{G}_F^{S_u}, D_{KL}(p_n||p_q) < \gamma_{KL}$, where $\gamma_{KL}$ is a similarity threshold set by the user on the ODVICE web interface and $D_{KL}(\cdot||\cdot)$ the KL divergence between distributions defined in the same probability space. A Monte Carlo sampling step (MCSample$(\cdot, \cdot)$) is then applied with sampling rate $\lambda_{KL}$ to decide how to sample nodes from the candidates discovered by this augmentation step. This entire procedure is applied recursively $\delta_H$ times to grow the graph nodes that are in direct proximity to $S_u$ by simply replacing $S_u$ with the expanded set of nodes produced by the previous iteration.

## ODVICE INTERFACE

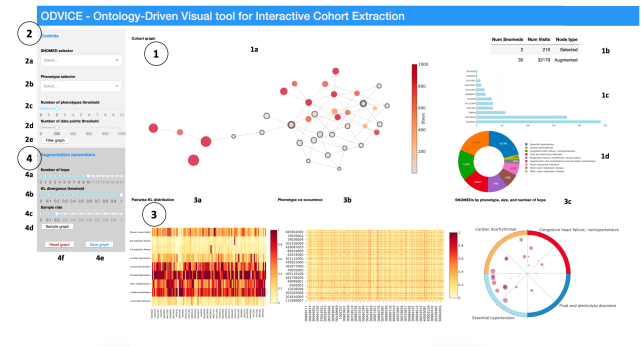

**Figure 1: The ODVICE web interface consists of four sections: (1) a primary visualization panel which displays the graphical structure of the data and summary information, (2) a control panel for filtering the graph, (3) a secondary visualization panel which displays information related to data purity and distribution, and (4) a panel for selecting the parameters of the augmentation algorithm.**

In this section, we describe a typical usage of ODVICE for an example analysis where the user is interested in an condition with limited patient records. The interface, as shown in Figure 1, consists of four sections: (1) A primary visualization panel that displays the observational data by overlaying the information over the ontological graph. It also displays the corresponding

**Algorithm 1:** Data Augmentation in ODVICE.

**Input** : filtered graph $\mathcal{G}_F$, specified nodes $S_u$, times of graph growth $\delta_H$, KL threshold $\gamma_{KL}$, sampling rate $\lambda_{KL}$

**Output**: augmented graph $\mathcal{G}_A$

1 $\mathcal{G}_A \leftarrow S_u$
2 $\tilde{S}_u \leftarrow \{n \in S_u : \text{descendants}_{\mathcal{G}_F}(n)\}$
3 **for** $i \leftarrow 1 : \delta_H$ **do**
4     sampledNodes = {}
5     **for** *each* $n \in \tilde{S}_u.keys()$ **do**
6         $N \leftarrow n.\text{getParents}_{\mathcal{G}_F}()$
7         **for** *each child* $\in \tilde{S}_u[n]$ **do**
8             $N' \leftarrow \text{child.getParents}_{\mathcal{G}_F}()$
9             $N = N \cup N'$
10     exclude nodes from $N \in \{n, \tilde{S}_u[n]\}$
11     **for** *each* $n_{canditate} \in N$ **do**
12         minKL $\leftarrow \infty$
13         **for** *each child* $\in \tilde{S}_u[n]$ **do**
14             minKL $\leftarrow$ min $\{D_{KL}(p_{n_{canditate}}||p_{child}), \text{minKL}\}$
15         **if** $minKL < \gamma_{KL}$ **then**
16             sampledNodes.append({$n_{canditate}$ : minKL})
17     selectedNodes = MCSample($\lambda_{KL}$, sampledNodes)
18     $\tilde{S}_u \leftarrow \{\text{selectedNodes} : \text{selectedNodes.descendants}_{\mathcal{G}_F}()\}$
19     $\mathcal{G}_A.\text{append}(\text{selectedNodes}, \text{descendants}_{\mathcal{G}_F}(\text{selectedNodes}))$
20 **return** $\mathcal{G}_A$

summary charts. (2) A control panel that allows the user to submit their diagnosis codes of interest as well as restrict the domain to phenotypes of interest. (3) A secondary visualization panel that shows more granular insights about selected records, with respect to both their phenotypical coverage as well their inter-nodes similarities. (4) A panel for specifying the parameters that triggers ODVICE to augment thier dataset.

Upon access to the ODVICE web interface, the user sees the primary visualization panel (1). This panel overlays the ontological graph (1a) derived from *SNOMED CT* showing the various medical concepts and their relationship. The color of each node indicates the number of records associated with that node, and the size is proportional to their position in the ontological hierarchy such that most specific nodes have the smallest size and the most general nodes have the biggest size. To the right, the interface displays summary statistics (1b) indicating the total number of records in the graph and the number of nodes in the displayed graph, a bar chart (1c) showing the distribution of the records over each node, and a pie chart (1d) exhibiting the coverage of each phenotype in the selected records.

The user can use the control panel (2) to specify their diagnosis of interest (2a) in terms of top-level SNOMED CT codes. They can also constrain the domain to the phenotypes of interest (2b) to generate a filtered subgraph from the original graph. They can further sepcify their choice by setting a threshold for the minimum number of phenotypes in each node of the filtered graph (2c), and a threshold for the minimum number of data points in each node of the filtered graph (2d).

**Table 1: Data sets generated by ODVICE for our experiments. $\eta_u$ and $\gamma_u^p$ were fixed to 100 and 200 respectively.**

| Data Set | $\gamma_{KL}$ | $\lambda_{KL}$ | $\delta_H$ |
|---|---|---|---|
| ODVICE 1 | 0.5 | 0.3 | 2 |
| ODVICE 2 | 0.4 | 0.2 | 2 |
| ODVICE 3 | 0.3 | 0.2 | 1 |

The secondary visualization panel (3) shows more granular information about the nodes of interest. This panel contains three sub-components that can provide the user insights about the inter-node similarity as well as phenotypical coverage that can be used to specify their preferences to select the augmented data.

The user by reviewing the presented information can, for example, specify the number of hops (4a) or to choose a low KL divergence threshold (4b) to constrain the algorithm to sample from similar distributions or adjust the sampling rate (4c). Finally, the user can then sample from the graph by applying the Monte Carlo sampling algorithm (4d) and save the graph for use in their subsequent analysis (4e) or reset the analysis back to the original graph (4f). Upon sampling, ODVICE updates the visualization panels and the ontological graph (1a) is updated to show the selected nodes with a thick gray border, their descendants with a thin gray border, and the augmented nodes with no borders.

## EXPERIMENTS

To evaluate the performance of ODVICE, we have focused on analytic studies for patients with fairly rare heart conditions reported in the MIMIC-III data set [7]. which is a large, single-center database with data on patients admitted to critical care units at a large tertiary hospital. We focused our experiments on vital signs, observations and diagnosis codes following the cohort construction proposed in [6]. Since MIMIC-III tracks diagnosis codes in the ICD-9 format, we have translated these codes into SNOMED CT using he one to one mapping [2] to construct the data graph $\mathcal{G}$ described in the Methods section. The two selected conditions forming our $S_u$ set of SNOMED CT codes are mitral valve stenosis with SNOMED CT code 79619009 and rheumatic disease of the heart valve with SNOMED CT code 16063004. We used the following set $P$ of phenotypes: Congestive heart failure; nonhypertensive, Cardiac dysrhythmias, Essential hypertension, Fluid and electrolyte disorders, Hypertension with complications and secondary hypertension, Acute myocardial infarction, Other lower respiratory disease, Other upper respiratory disease, Respiratory failure; insufficiency; arrest (adult). The resulting ontology graph $\mathcal{G}$ consisted of 9118 SNOMED codes mapping into 41780 ICU visits for 33569 unique patients.

We applied ODVICE to solve two classification tasks:

(1) **In Hospital Mortality Prediction**: Using the first 48 hours of data from patient visits, we trained ML models to predict whether the patient would die within the remaining part of her hospital stay. We excluded all visits with less than 48 hours of data from this analysis.

---
[2]https://www.nlm.nih.gov/research/umls/mapping_projects/icd9cm_to_snomedct.html

(2) **Myocardial Infarction Phenotyping**: Using all the data for each patient visits, we developed ML models attempting to correctly guess the myocardial infarction phenotype.

We compared ODVICE against 4 baseline strategies for data collection and augmentation:

- **Target:** Using only the visits from the nodes in $S_u$ with one of the target phenotypes mentioned before.
- **Random 1:** Augmenting **Target** with data randomly selected from $\mathcal{G}_F$ to produce a data set with 3000 visits.
- **Random 2:** Augmenting **Target** with data randomly selected from $\mathcal{G}_F$ to produce a data set with 6000 visits.
- **Random 3:** Augmenting **Target** with data randomly selected from $\mathcal{G}_F$ to produce a data set with 8000 visits.

We also tried various configuration of ODVICE in our experiments. Table 1 shows the set of parameters used to evaluate ODVICE. We tried 3 different ODVICE augmentation strategies. ODVICE 1 produced the largest amount of augmented data as it used the highest values for $\gamma_{KL}$, $\lambda_{KL}$ and $\delta_H$. ODVICE 2 was less aggressive. ODVICE 3 produced the least amounts augmented of data and was hence more constrained to the phenotypes and SNOMED CT code used. For each of these tasks, we tried two common machine learning algorithms: Logistic Regression (LR) and Random Forest (RF). Each learning algorithm was applied using a 3 fold cross validation approach. The performance was measured using the Area Under the ROC curve (AUC).

Table 2 shows AUC results obtained for all baselines and ODVICE for the **In Hospital Mortality Prediction** and the **Myocardial Infarction Phenotyping** tasks. In both cases, we observed a significant AUC gain obtained by applying ODVICE for these tasks when compared with all our baselines. Furthermore, we also observed that the ODVICE 3 data set that has data very similar to target nodes (with a number of hops $\delta_H$ set to 1) tend to outperform the other ODVICE data augmentation strategies, thus hinting at not only the importance of carefully constructing cohorts for predictive tasks but also hinting at the effectiveness of guiding this cohort construction using medical concepts that are semantically close to the target data set.

## CONCLUSION

We presented ODVICE, a framework for ontology-driven data augmentation that enables end users to interactively augment their data set with a visual analytic tool. ODVICE accomplishes this using a novel Monte-Carlo graph-based algorithm that combines both ontological knowledge and observational evidence to suggest the most relevant records for data augmentation. We tested the performance of the method on real-world cohort construction problems for prediction and computational phenotyping. The experimental results presented demonstrate the effectiveness of the proposed method. In the future, we plan on testing ODVICE's cohort construction capabilities on more computational health problems and on larger general EHR data sets.

Table 2: Comparison of different data augmentation strategies

| Data Set | The "in hospital mortality" prediction task | | | The "myocardial infarction" phenotyping task | | |
|---|---|---|---|---|---|---|
| | Data Description | AUC for LR | AUC for RF | Data Description | AUC for LR | AUC for RF |
| Target only | 131 visits (14 +, 117 - ) | 0.64(0.17) | 0.70(0.08) | 210 visits (15 +, 195 -) | 0.44(0.06) | 0.66(0.02) |
| Random 1 | 3000 visits (417 +, 2583 -) | 0.66(0.05) | 0.73(0.02) | 3000 visits (321 +, 2679 -) | 0.47(0.09) | 0.51(0.05) |
| Random 2 | 6000 visits (1788 +, 5212 -) | 0.78(0.02) | 0.78(0.02) | 6000 visits (640 +, 5360 -) | 0.56(0.01) | 0.61(0.03) |
| Random 3 | 8000 visits (1046 +, 6954 -) | 0.78(0.01) | 0.82(0.01) | 8000 visits (849 +, 7151 -) | 0.46(0.01) | 0.52(0.07) |
| ODVICE 1 | 11027 visits (1888 +, 9139 -) | 0.80(0.05) | 0.77(0.05) | 18188 visits (1605 +, 16583 -) | 0.47(0.02) | 0.63(0.05) |
| ODVICE 2 | 7890 visits (1165 +, 6725 -) | **0.83(0.00)** | 0.81(0.00) | 12908 visits (2165 +, 10743 -) | 0.52(0.02) | 0.67(0.01) |
| ODVICE 3 | 1856 visits (180 +, 1676-) | 0.81(0.05) | **0.91(0.05)** | 3250 visits (555 +, 2695 -) | **0.72(0.00)** | **0.87(0.04)** |